\documentclass{ecai} 

\usepackage{latexsym}
\usepackage{amssymb}
\usepackage{amsmath}
\usepackage{amsthm}
\usepackage{booktabs}
\usepackage{enumitem}
\usepackage{graphicx}
\usepackage{color}
\usepackage{algorithm}
\usepackage{algorithmic}
\usepackage[utf8]{inputenc}
\usepackage{times}
\usepackage[hidelinks]{hyperref}
\usepackage{url}
\usepackage{caption}
\usepackage{orcidlink}

\newcommand{\BibTeX}{B\kern-.05em{\sc i\kern-.025em b}\kern-.08em\TeX}

\begin{document}

%%%%%%%%%%%%%%%%%%%%%%%%%%%%%%%%%%%%%%%%%%%%%%%%%%%%%%%%%%%%%%%%%%%%%%%%

\begin{frontmatter}

%%% Use this command to specify your submission number.
%%% In doubleblind mode, it will be printed on the first page.

\paperid{8362} 

%%% Use this command to specify the title of your paper.

\title{Full-History Graphs with Edge-Type Decoupled Networks for Temporal Reasoning}

%%% Use this combinations of commands to specify all authors of your 
%%% paper. Use \fnms{} and \snm{} to indicate everyone's first names 
%%% and surname. This will help the publisher with indexing the 
%%% proceedings. Please use a reasonable approximation in case your 
%%% name does not neatly split into "first names" and "surname".
%%% Specifying your ORCID digital identifier is optional. 
%%% Use the \thanks{} command to indicate one or more corresponding 
%%% authors and their email address(es). If so desired, you can specify
%%% author contributions using the \footnote{} command.

\author[A]{\fnms{Osama}~\snm{Mohammed}\thanks{Corresponding Author. Email: osama.mohammed@ki.uni-stuttgart.de}\orcidlink{0009-0002-3228-3036}}
\author[A]{\fnms{Jiaxin}~\snm{Pan}\orcidlink{0000-0003-1055-7104}}
\author[A]{\fnms{Mojtaba}~\snm{Nayyeri}\orcidlink{0000-0002-9177-0312}} 
\author[A]{\fnms{Daniel} \snm{Hernández}\orcidlink{0000-0002-7896-0875}} 
\author[A,B]{\fnms{Steffen}~\snm{Staab}\orcidlink{0000-0002-0780-4154}} 

\address[A]{Institute for Artificial Intelligence, University of Stuttgart, Germany}
\address[B]{University of Southampton, UK}

%%% Use this environment to include an abstract of your paper.

\begin{abstract}
Modeling evolving interactions among entities is critical in many real-world tasks. 
For example, predicting driver maneuvers in traffic requires tracking how neighboring vehicles accelerate, brake, and change lanes relative to one another over consecutive frames. 
Similarly, detecting financial fraud hinges on following the flow of funds through successive transactions as they propagate across the network. 
Unlike classic time-series forecasting, these settings demand reasoning over \emph{who} interacts with \emph{whom} and \emph{when}, calling for a \textit{temporal-graph representation} that makes both the relations and their evolution explicit. Existing temporal-graph methods use snapshot graphs to represent temporal evolution. In this paper, we introduce a \emph{full-history graph} that instantiates one node for every entity at every timestep and separates two edge sets: (i) \textit{intra-timestep} edges that capture relations within a single frame, and (ii) \textit{inter-timestep} edges that connect an entity to itself at consecutive steps. To learn on this graph we design an \textbf{Edge-Type Decoupled Network} (ETDNet) with parallel modules: a graph-attention module aggregates information along intra-timestep edges, a multi-head temporal-attention module attends over an entity’s inter-timestep history, and a fusion module combines the two messages after every layer. When evaluated on driver-intention prediction (Waymo) and Bitcoin fraud detection (Elliptic++), ETDNet consistently surpasses strong baselines, lifting Waymo joint accuracy to 75.6 \% (vs.\ 74.1 \%) and raising Elliptic++ illicit-class F1 to 88.1 \% (vs.\ 60.4 \%). 
These gains demonstrate the benefit of representing structural and temporal relations as distinct edges in a single graph.
\end{abstract}

\end{frontmatter}

%%%%%%%%%%%%%%%%%%%%%%%%%%%%%%%%%%%%%%%%%%%%%%%%%%%%%%%%%%%%%%%%%%%%%%%%

\section{Introduction}

\begin{figure}[t]   
    \centering    \includegraphics[width=0.45\textwidth]{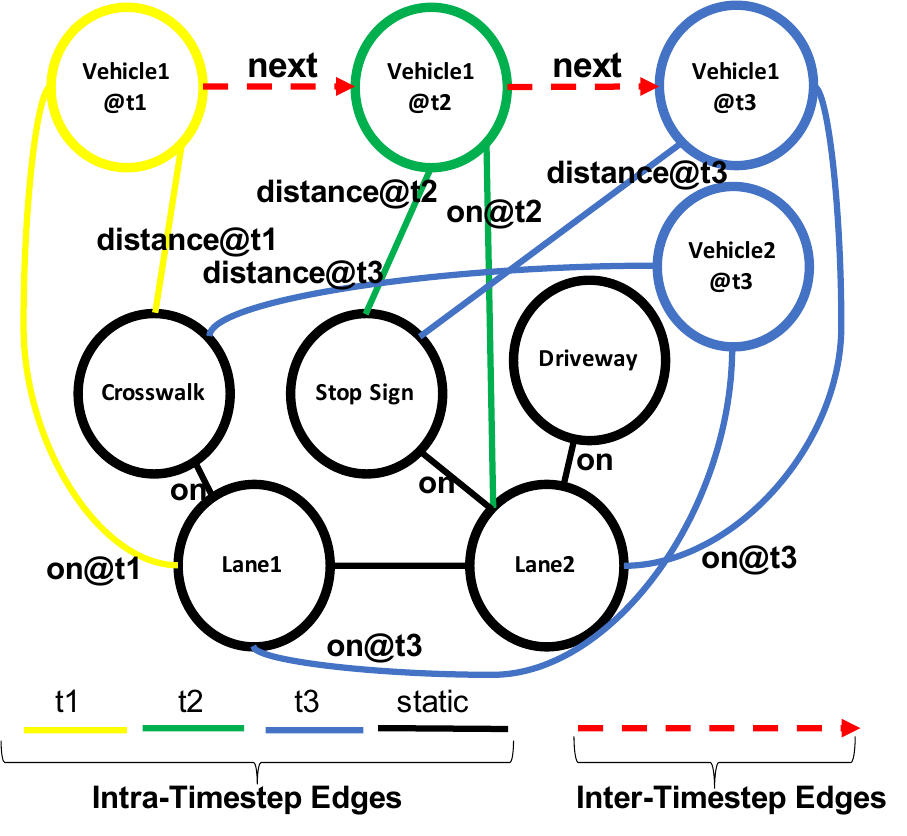}
    \caption{Example full-history graph with three timesteps (\(t{=}1,2,3\)).
         Dynamic entity \textit{Vehicle 1} appears as three nodes
         linked by dashed red temporal edges; static road objects
         interact with it in each timestep.}
    \label{fig:fhg_intro}
    \vspace{0.5cm}
\end{figure}

Forecasting driver maneuvers in traffic systems requires reasoning about how cars, lanes, and signals influence one another from frame to frame.  
Financial fraud detection likewise hinges on tracing the flow of funds through successive transactions.  
In both settings, the core difficulty is to capture \emph{instantaneous relations}: who is related to whom, who pays whom, and how those relations evolve over time.

Conventional time sequence models (e.g., RNNs or Transformers run on a single time-series) process each entity’s record and therefore fail to exploit those cross-entity links.  
A \textbf{temporal-graph} representation, by contrast, keeps both the interaction structure and the timing of each event explicit.

Prior work in temporal GNN (TGNN) follows two complementary strategies. \emph{Snapshot methods} unroll a dynamic graph into a sequence of static snapshots, apply a GNN (e.g., GCN or GAT) at every timestep, and then feed the per-snapshot embeddings to a recurrent or transformer temporal aggregator \cite{longa2023graph,pareja2020evolvegcn,wang2024novel,xie2023gtea}. \emph{Integrated methods} embed time directly in the graph computation; either through continuous-time event processes that update node states at every interaction \cite{rossi2020tgn,wen2022trend,trivedi2019dyrep} or through attention layers that assign learnable, recency-aware weights to edges \cite{xu2020tgat,sasal2024tempokgat,xie2023targat}. Because snapshot methods \emph{aggregate} events across multiple fine-grained timesteps and truncate the sequence they process, several studies report that they \emph{tend} to lose fine-grained ordering and long-range temporal cues \cite{longa2023graph,chen2023neutronstream,bravo2024truncation}. Likewise, integrated models that mix structural and temporal signals in a single aggregation step risk blurring the two factors and weakening long-horizon reasoning; theoretical and empirical analyses link this to over-smoothing and over-squashing of temporal information \cite{sankar2020dysat,beddarwiesing2024wl_dynamic,longa2023graph}. For example, in traffic forecasting, such blurring can hide multi-step manoeuvre cues, while in fraud detection, it can obscure multi-hop laundering paths that resurface after long dormancies.

\begin{figure}[t]   
    \centering    \includegraphics[width=0.45\textwidth]{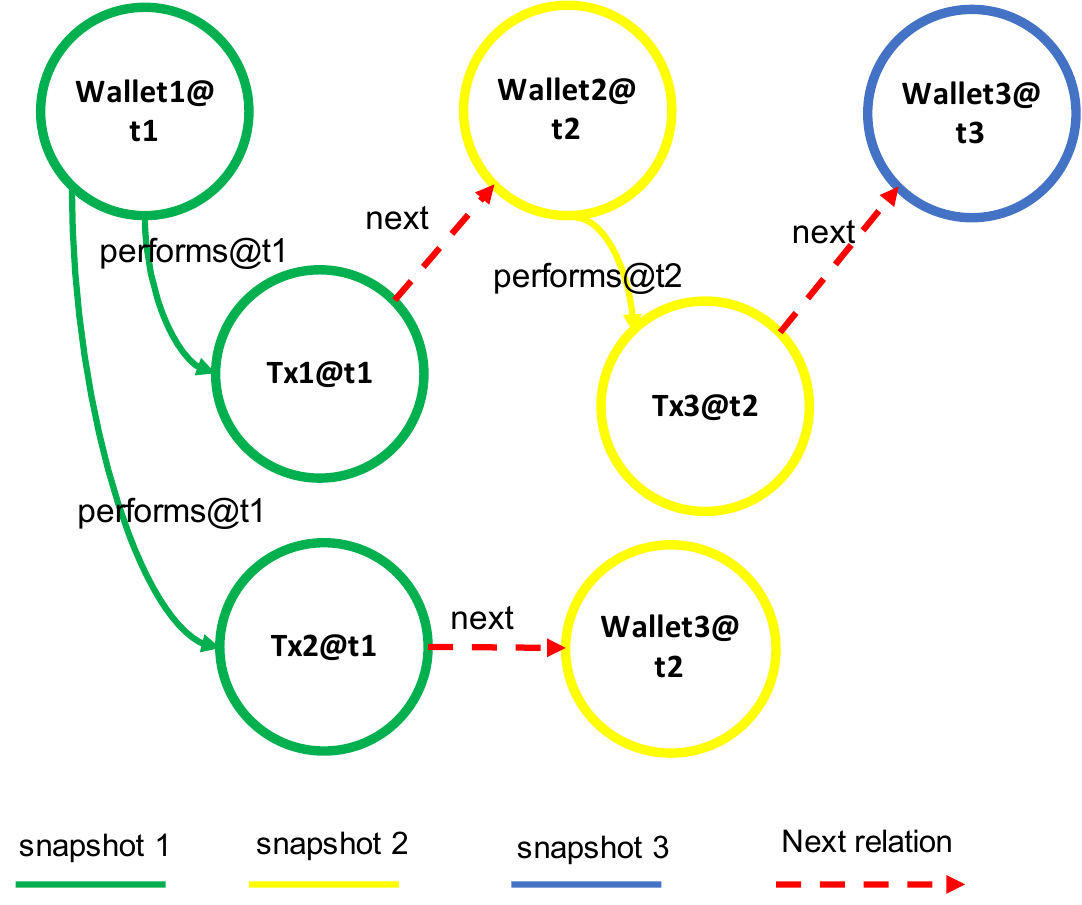}
    \caption{Example full-history graph with three timesteps (\(t{=}1,2,3\)).
         Dynamic entity such as wallets and transactions (tx) appears as different nodes
         across different timesteps linked by dashed red temporal edges.}
    \label{fig:fhg_intro_elliptic}
    \vspace{0.5cm}
\end{figure}

To overcome these limitations, we propose a representation consisting of a single graph that encodes the \emph{entire} observation history.
Figure~\ref{fig:fhg_intro} shows a traffic scene example, but the same representation approach likewise applies to other dynamic domains (e.g., financial-transaction networks).  
For every entity \(u\) observed at timestep \(t\), we create a node \(u^{t}\) (e.g., the three colored nodes for \textit{Vehicle 1} in Figure~\ref{fig:fhg_intro}); we call the resulting time-unfolded structure a \textbf{full-history graph (FHG)}.
Two disjoint edge sets make semantics explicit:

\begin{itemize}[leftmargin=1.2em,topsep=5pt,itemsep=3pt,parsep=0pt]
  \item \textbf{Intra-timestep edges} (solid green, yellow, and blue in
        Figure~\ref{fig:fhg_intro}) link entities that coexist in the same
        timestep, capturing instantaneous interactions.
  \item \textbf{Inter-timestep edges} (dashed red) connect \emph{temporally
        successive events}, most commonly the same entity at \(t\) and
        \(t{+}1\), but also cross-entity hand-offs such as a transaction chain, thereby encoding the flow of time, as can be seen in Figure~\ref{fig:fhg_intro_elliptic}.
\end{itemize}

Full history graphs may be applied directly to any domain where events unfold in ordered steps.
It preserves both fine-grained temporal ordering and structural context,
and it serves as the backbone for our novel dual-branch GNN introduced in
Section~\ref{sec:method}.

Our model adopts a parallel message passing strategy. At each GNN layer, one round of message passing occurs over instantaneous relationships first, while at the same time, another round of message passing over history edges occurs. The first branch leverages a graph attention block to model intra-timestep relationships among nodes. While effective at capturing intra-timestep relationships, GNNs are prone to over-smoothing, especially in deeper layers. Therefore, the second branch integrates a self-attention-based block that processes inter-timestep edges and preserves feature diversity across time. By combining these branches, our model captures both immediate context and long-range dependencies.

We demonstrate the effectiveness of our approach on two challenging real-world applications: driver-intention forecasting and dynamic fraud detection in transaction networks. Our experimental results show that our model outperforms existing baselines by leveraging the full temporal history of the data.

Our main contributions are:
\begin{enumerate}[leftmargin=*]
    \item A full-history graph representation with two edge sets: (i) intra-timestep edges connecting entities that interact within the same frame, and (ii) inter-timestep edges linking temporally successive events through self-links or cross-entity hand-offs.
    \item A dual-branch network consisting of (i) graph attention for intra-timestep edges and (ii) temporal self-attention for inter-timestep edges, thereby avoiding the depth limits and over-smoothing issues of GNNs.
    \item Extensive experiments across traffic and finance demonstrating superior accuracy, ablations confirming each design choice, and analyses of runtime \& memory.
\end{enumerate}

The remainder of this paper is organized as follows. Section~\ref{sec:related} reviews related work. Section~\ref{sec:method} describes our full-history graph construction and dual-branch neural architecture. Section~\ref{sec:experiments} presents experimental results, and Section~\ref{sec:conclusion} concludes with discussions on future directions.

%%%%%%%%%%%%%%%%%%%%%%%%%%%%%%%%%%%%%%%%%%%%%%%%%%%%%%%%%%%%%%%%%%%%%%%%

\section{Related Work}

\label{sec:related}
\begin{figure*}[t]
    \centering
    \includegraphics[width=\textwidth,height=0.2\textheight,keepaspectratio]{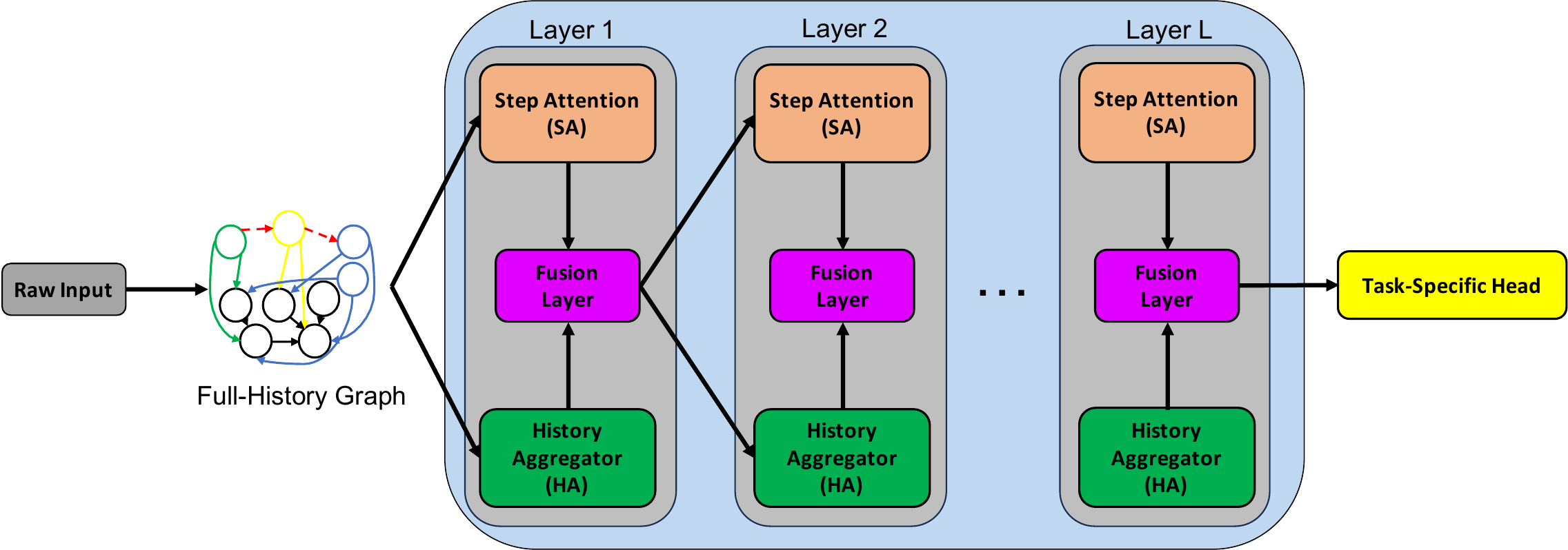}
    \caption{End-to-end \textbf{ETDNet} encoder. Raw data are converted to a \textit{full-history graph}; each layer then applies \emph{Step Attention} (\textbf{SA}) on domain edges and \emph{History Attention} (\textbf{HA}) on temporal edges. A residual \emph{Fusion Layer} merges the output of the two messages and feeds the next layer. After $L$ layers, a lightweight \emph{task-specific head} produces the final predictions.}
    \label{fig:dualbranch_agg}
    \vspace{0.5em}
\end{figure*}

Temporal graph neural networks (TGNNs) adapt static GNNs to evolving relational data, enabling applications in recommendation, traffic forecasting, and blockchain analytics~\cite{longa2023graphsurvey}.  Existing methods can be grouped into three lines of research:  
(i) \textit{memory–based} models that combine a GNN with recurrent or external-memory updates,  
(ii) \textit{time-aware attention} models that extend Transformer or GAT layers with temporal encodings, and  
(iii) \textit{decoupled} architectures that process spatial and temporal edges in separate branches before fusing the messages.  
The proposed ETDNet falls into the third group but differs by operating on a full-history graph and by assigning a dedicated attention mechanism to each edge family, yielding clearer inductive bias and lower memory than prior decoupled designs.
%------------------------------------------------------------%
\paragraph{Snapshot GNN baselines.}
Early works treat a dynamic graph as a sequence of static “snapshots’’
and run a conventional GNN at every step, optionally sharing weights.
Representative baselines include GCN~\cite{kipf2017semisupervised},
Graph\-SAGE~\cite{hamilton2017inductive}, and the spatio-temporal
extension ST-GCN~\cite{yu2017spatio}.  
While effective on short horizons, snapshot models conflate spatial and
temporal signals inside the same message‐passing layer and must stack
many layers to reach distant timesteps, which amplifies
over-smoothing.  
Our experiments (Section \ref{sec:experiments}) confirm that performance drops
once trajectories exceed two–three hops, motivating an explicit
temporal branch.

\paragraph{Memory–bank temporal GNNs.}
To avoid recomputing snapshots, memory-based TGNNs maintain a per-node
state that is updated whenever an event arrives.  TGN stores incoming
messages in a mailbox and applies a GRU update before each
prediction~\cite{rossi2020tgn}; JoDIE augments this idea with coupled
recurrent units for users and items~\cite{kumar2019predicting}.  
Memory banks capture long histories but require $O(|V|H)$ storage
(nodes $\times$ window) and mix spatial and temporal information in the
same recurrent cell.  ETDNet discards the global mailbox and instead
attends only over the \emph{local} window of $\mathcal{H}$ predecessors,
cutting memory while still preserving long temporal context.

\paragraph{Continuous-time attention kernels.}
A second line of work replaces discrete updates with continuous-time
encodings.  TGAT introduces sinusoidal time
features~\cite{xu2020tgat}; DyRep uses Hawkes processes to model event
intensity~\cite{trivedi2019dyrep}; DyGFormer couples these ideas with a
global transformer and relative-time biases~\cite{huang2023tgb}.
Such kernels improve temporal fidelity but still merge spatial and
temporal neighbors inside a single attention matrix, leaving
edge-type–specific inductive bias unexplored.  ETDNet keeps the
continuous-time benefits (windowed self-attention) yet processes
domain $D$ and temporal $\mathcal{H}$ edges in \emph{distinct}
modules, allowing each branch to specialize.

\paragraph{Spatial–temporal separation.}
Recent work has begun to decouple the two channels.
STGNN applies separate convolutions to static and dynamic
adjacency \cite{10.14778/3551793.3551827}; TG-NeXt factorizes message passing by
edge timestamp and direction~\cite{wu-etal-2020-temp};
SemanticFormer stacks two semantic “heads’’ for actor and map graphs in
trajectory forecasting \cite{sun2024semanticformer}.
These models validate the benefit of separation but either rely on
hand-crafted channels (STGNN, SemanticFormer) or keep a snapshot
backbone (TG-NeXt).  
Our contribution complements them with (i) a \emph{full-history} graph
representation that makes temporal causality explicit and
(ii) a dual-attention layer (SA + HA) that assigns dedicated parameters
to each edge family while remaining lightweight.

In summary, prior TGNNs leave three gaps.  
(i) \textit{Snapshot} stacks conflate space–time signals and over-smooth after a few hops~\cite{yu2017spatio,longa2023graph,chen2023neutronstream}.  
(ii) \textit{Memory–bank} or continuous-time models retain history but still mix edge types and incur \(O(|V|B)\) storage~\cite{rossi2020tgn,xu2020tgat,trivedi2019dyrep,huang2023tgb}.  
(iii) Recent “decoupled’’ networks use handcrafted channels or snapshot backbones~\cite{10.14778/3551793.3551827,wu-etal-2020-temp,sun2024semanticformer}.  
Our ETDNet fills these gaps with a \emph{full-history} graph that keeps causal order explicit and a lightweight \emph{dual-attention} layer that treats intra- and inter-timestep edges separately, yielding lower memory and stronger long-range reasoning.

%%%%%%%%%%%%%%%%%%%%%%%%%%%%%%%%%%%%%%%%%%%%%%%%%%%%%%%%%%%%%%%%%%%%%%%%

\section{Methodology}

\label{sec:method}

We propose a full-history graph learning framework that encapsulates time-evolving entities and their interactions in a unified structure, followed by learning from this structure. Section~\ref{subsec:graph_repr} formalizes the graph; Section~\ref{subsec:dual_branch} presents ETDNet; Section~\ref{subsec:training} details training.

%%%%%%%%%%%%%%%%%%%%%%%%%%%%%%%%%%%%%%%%%%%%%%

\subsection{Full-History Graph Representation}
\label{subsec:graph_repr}

Let \(T=\{0,1,\dots,\tau_{\max}\}\) be the discrete set of timesteps.  
A full-history graph is a graph \(G=(\widetilde V,E)\) whose node set \(\widetilde V\) is the union of two disjoint families:
\emph{dynamic nodes}, denoted by \(U\), and \emph{static nodes}, denoted by \(S\).
Dynamic nodes correspond to entities that evolve over time (e.g., vehicles, transactions), while static nodes represent entities fixed in place (e.g., stop signs).  

For each dynamic entity \(u\in U\) and each timestep \(t\in T_u\subseteq T\) in which \(u\) is present,  
we create a replica node \(u^{t}\) with feature vector \(\mathbf{x}_{u,t}\in\mathbb{R}^{d}\).  
Every static entity \(s\in S\) is represented by a single node \(s\) whose feature \(\mathbf{x}_{s}\) is shared across all timesteps.  
The complete node set is
\[
\widetilde V
\;=\;
\underbrace{\{\,u^{t}\mid u\in U,\;t\in T_u\,\}}_{\text{dynamic nodes}}
\;\cup\;
\underbrace{S}_{\text{static nodes}} .
\]

\paragraph{Edge types.}
For each timestep \(t\), let
\(C_t := \{u^{t}\mid u\in U,\; t\in T_u\}\;\cup\; S\) be the carrier set that includes the dynamic replicas present at \(t\) and all static nodes.  
We distinguish two types of edges:

\begin{enumerate}[label=(\roman*),leftmargin=*]
  \item \textbf{Intra-timestep edges}  
        \(D \subseteq \bigcup_{t\in T} C_t \times C_t\) link entities that interact \emph{within} timestep \(t\).
  \item \textbf{Inter-timestep edges}
        \(\mathcal{H} \subseteq \bigcup_{t<\tau_{\max}} C_t \times C_{t+1}\) connect \emph{temporally successive events}.
        The most common case is the self-edge \((u^{t},u^{t+1})\), but \(\mathcal{H}\) can also include cross-entity links \((p^{t},\,q^{\,t+1})\), such as when a transaction at \(t\) initiates a related one at \(t{+}1\).
\end{enumerate}

The full-history graph is then
\[
G = (\widetilde V,\,E), 
\qquad 
E = D \cup \mathcal{H}.
\]

This construction (i) retains the full temporal record (no sliding windows), (ii) lets us process $D$ and $\mathcal{H}$ with separate modules (Sec.~\ref{subsec:dual_branch}), and (iii) accommodates static entities and cross-entity temporal flows, making the template transferable from traffic to transactions. Because every $\mathcal{H}$ edge points $t\!\rightarrow\!t{+}1$, paths along $\mathcal{H}$ are acyclic, while $D$ stays within one timestep and may form cycles; keeping the two edge families separate preserves temporal order and enables distinct updates for simultaneous vs.\ cross-time interactions.

Figures~\ref{fig:fhg_intro} and \ref{fig:fhg_intro_elliptic} illustrate a three-timestep traffic scenario and Bitcoin transactions, respectively: colored solid edges show intra-timestep interactions, while dashed red edges mark the inter-timestep \emph{next} links.

%%%%%%%%%%%%%%%%%%%%%%%%%%%%%%%%%%%%%%%%%%%%%%
\subsection{Edge-Type Decoupled Architecture}
\label{subsec:dual_branch}

\begin{figure}[t]
    \centering
    \includegraphics[width=0.85\columnwidth]{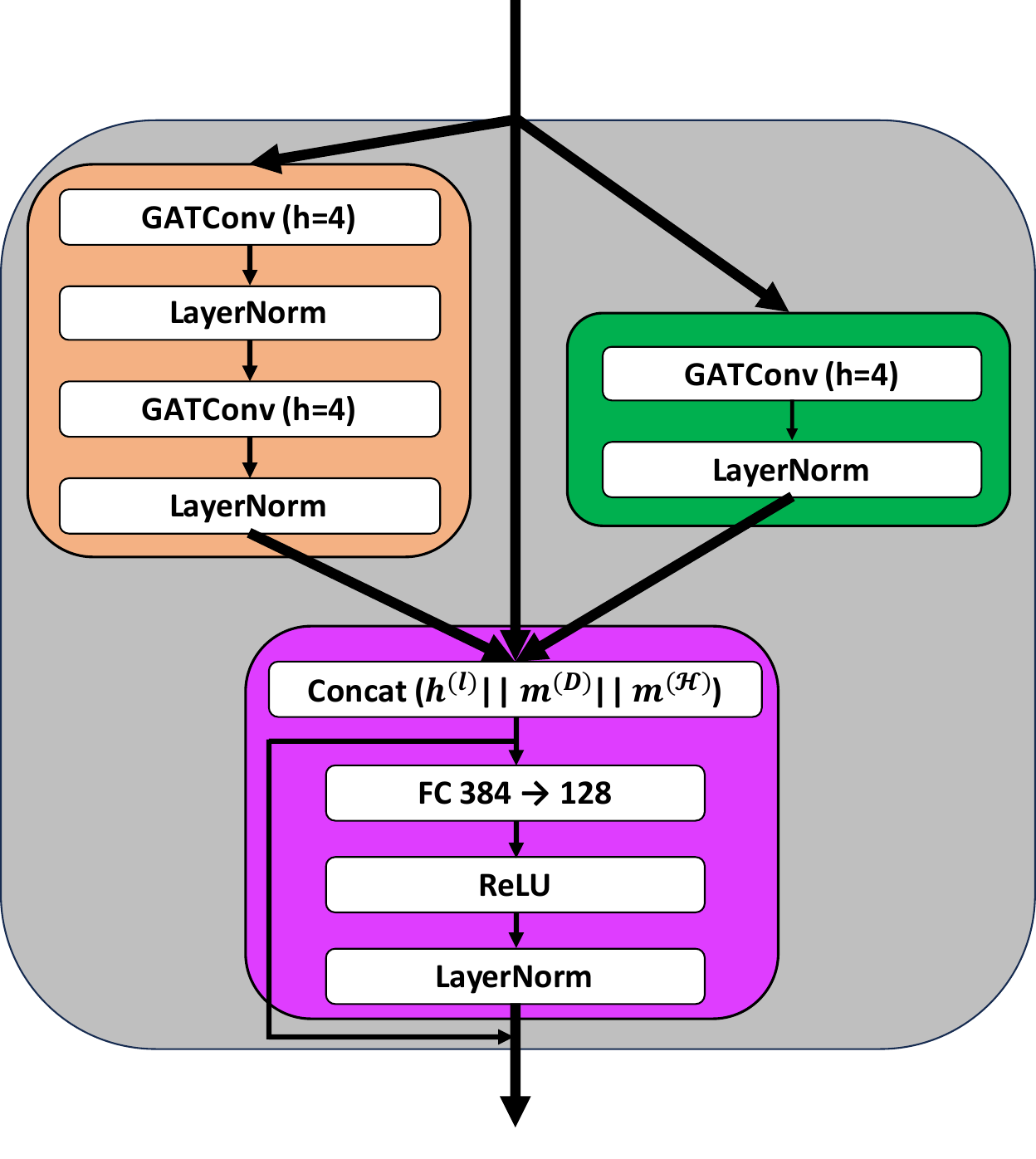}
    \caption{Inside one ETD layer. \textbf{(left)} SA stack: \(K_s\) GATConv sublayers, each with \(H_s\) heads, followed by LayerNorm and a residual arrow. \textbf{(right)} HA block: \(H_t\)-head multi-head attention over the \(B\)-length window, followed by LayerNorm. \textbf{(bottom)} Fusion Layer concatenates \(h^{(\ell)}\), \(m^{D}\), and \(m^{\mathcal{H}}\), applies a \(3d\!\rightarrow\! d\) linear projection, ReLU, and LayerNorm, then outputs \(h^{(\ell+1)}\). Arrows show residual connections.}

    \label{fig:etd_layer_zoom}
    \vspace{3em}
\end{figure}

After we construct the full-history graph, we use our proposed network, Edge-Type Decoupled Network (ETDNet), to learn on the graph. ETDNet takes as input the full-history graph and stacks $L$ identical layers. Within each layer, ETDNet aggregates intra-timestep context, attends over recent history, and fuses the two with the previous embedding; after the final layer, embeddings go to a task head. Figure~\ref{fig:dualbranch_agg} presents the overall ETDNet architecture, while Figure~\ref{fig:etd_layer_zoom} zooms into a single ETD layer.

\paragraph{Block overview.} Each layer contains three blocks:
\begin{enumerate}[label=\arabic*.,leftmargin=*]
\item \textbf{Step Attention (SA)} A stack of $K_s$ sublayers, each containing $H_s$ attention heads, applied over $D$
\item \textbf{History Attention (HA)} A self-attention block with $H_t$ attention heads over the past \(B\) steps of \(\mathcal{H}\);
\item \textbf{Fusion Layer (FL)} a residual MLP that merges the current embedding with the two messages.
\end{enumerate}

Unlike many snapshot GNNs, HA therefore sees a \emph{set} of time-ordered predecessors, not just the node’s own past, and degenerates to a simple self-chain when $\mathcal{H}$ contains only
\((u^{t-1},u^{t})\) edges.

Below we denote the embedding of node $x$ at layer $l$ by
$h^{(l)}_{x}\!\in\!\mathbb{R}^{d}$ ($h^{(0)}_{x}=\mathbf{x}_{x}$).
We update only dynamic replicas $u^{t}$; static-node features $\mathbf{x}_{s}$ serve as neighbors in $D$ but remain fixed across layers.

\subsubsection*{1.\;Step Attention (SA)}

SA aggregates information over \emph{intra-timestep} edges \(D\).
For each node \(u^{t}\) we set
\(\mathcal{N}_{D}(u^{t})=\{\,v^{t}\mid(v^{t},u^{t})\!\in\!D\}\).
Inside SA we stack \(K_{s}\) identical sublayers; every sublayer
uses \(H_{s}\) attention heads. We set \(h^{(l,0)}_{u^{t}} := h^{(l)}_{u^{t}}\) and update through the \(K_s\) sublayers to \(h^{(l,K_s)}_{u^{t}}\).

\smallskip
With head index \(r\) and sublayer index \(k\) we first compute attention scores
\begin{equation}
e_{uv}^{(r,k)} =
\mathrm{LeakyReLU}\!\Bigl(
\mathbf{a}_{r,k}^{\!\top}
\bigl[\,\mathbf{W}_{Q,r,k}\,h^{(l,k-1)}_{u^{t}}
\;\Vert\;
\mathbf{W}_{K,r,k}\,h^{(l,k-1)}_{v^{t}}\,\bigr]\Bigr),
\label{eq:sa-score}
\end{equation}
where  
\(\mathbf{W}_{Q,r,k},\mathbf{W}_{K,r,k}\!\in\!\mathbb{R}^{d'\times d}\)
project queries and keys to the per-head width \(d' = d/H_s\) (with $d$ the model dimension), and \(\mathbf{a}_{r,k}\!\in\!\mathbb{R}^{2d'}\) is the learnable
scoring vector.
We then normalize the attention score, as
\begin{equation}
\alpha_{uv}^{(r,k)} =
\frac{\exp(e_{uv}^{(r,k)})}{
\sum_{w^{t}\in\mathcal{N}_{D}(u^{t})}\exp(e_{uw}^{(r,k)})},
\label{eq:sa-alpha}
\end{equation}
thus, \(\alpha_{uv}^{(r,k)}\) is the attention weight of neighbor
\(v^{t}\) with respect to \(u^{t}\) for head \(r\).
We then perform neighborhood aggregation by
\begin{equation}
m_{u^{t}}^{(r,k)} =
\sum_{v^{t}\in\mathcal{N}_{D}(u^{t})}
\alpha_{uv}^{(r,k)}\;
\mathbf{W}_{V,r,k}\,h^{(l,k-1)}_{v^{t}},
\label{eq:sa-agg}
\end{equation}
where  
\(\mathbf{W}_{V,r,k}\!\in\!\mathbb{R}^{d'\times d}\) maps values to the head space.

\smallskip

After that, we compute the residual update. 
All $H_s$ head messages are concatenated, projected back to \(d\),
added to the residual, and normalized:
\begin{equation}
h^{(l,k)}_{u^{t}} =
\mathrm{LayerNorm}\!\Bigl(
\mathbf{O}_{k}\,[m_{u^{t}}^{(1,k)}\Vert\dots\Vert m_{u^{t}}^{(H_s,k)}]
+ h^{(l,k-1)}_{u^{t}}\Bigr),
\label{eq:sa-res}
\end{equation}
with shared output projection
\(\mathbf{O}_{k}\!\in\!\mathbb{R}^{d\times d}\).

\smallskip
Stacking the $K_s$ SA sublayers gives a $K_s$-hop receptive field restricted
to the current timestep while keeping the network shallow.  The final intra-timestep message is
\(m^{D}_{u^{t}} := h^{(l,K_s)}_{u^{t}}\), which is passed to the Fusion
Layer.

\subsubsection*{2.\;History Attention (HA)}

HA aggregates information that arrives \emph{through the inter-timestep
edge set} $\mathcal{H}$.  For the current node $u^{t}$ we collect up to
$B$ predecessors reachable along $\mathcal{H}$:
\[
\mathcal{P}_{B}(u^{t})
=\Bigl\{\,w^{\tau}\;\Big|\;
       w^{\tau}\!\in\!\operatorname{Hop}_{\le B}(\mathcal{H},u^{t}),\;
       \tau<t\Bigr\},
\]
where $\operatorname{Hop}_{\le B}(\mathcal{H},u^{t})$ is the set of nodes
that can reach $u^{t}$ via a directed path of length \(\le B\) in
$\mathcal{H}$.  In a driver-intention graph this set contains exactly
$u^{t-1}$ (self-edge); in a transaction graph it may contain several
transfers that “fan in’’ to the current one.  We order
$\mathcal{P}_{B}(u^{t})$ by time, obtain \(m=\lvert\mathcal{P}_{B}(u^{t})\rvert\le B\),
and zero-pad to length $B$:
\[
Z_{u,t} =
\bigl[h^{(l)}_{w^{\tau_1}}\;\Vert\;
      \dots\;\Vert\;
      h^{(l)}_{w^{\tau_m}}\bigr]\in\mathbb{R}^{B\times d}.
\]

Using head index $r\in\{1,\dots,H_{t}\}$ we compute the query–key–value
projections
\begin{equation}
Q_{r}=Z_{u,t}\mathbf{W}_{Q,r}^{\top},\quad
K_{r}=Z_{u,t}\mathbf{W}_{K,r}^{\top},\quad
V_{r}=Z_{u,t}\mathbf{W}_{V,r}^{\top},
\label{eq:ha-qkv}
\end{equation}
where
$\mathbf{W}_{Q,r},\mathbf{W}_{K,r},\mathbf{W}_{V,r}\!\in\!\mathbb{R}^{d''\times d}$
are shared, learnable matrices and $d''=d/H_{t}$.

We then compute the scaled dot-product attention
\begin{equation}
A_{r}=\operatorname{Softmax}\!\bigl(Q_{r}K_{r}^{\!\top}/\sqrt{d''}\bigr),\qquad
O_{r}=A_{r}V_{r},
\label{eq:ha-attn}
\end{equation}
where $A_{r}\!\in\!\mathbb{R}^{B\times B}$ and
$O_{r}\!\in\!\mathbb{R}^{B\times d''}$.

Finally, we take the most-recent row $O_{r}[m-1,:]$ from each attention
head, concatenate them, project back to dimension \(d\), and apply layer
normalization:
\begin{equation}
m^{\mathcal{H}}_{u^{t}}=
\operatorname{LayerNorm}\!\Bigl(
\Bigl[\!\bigl\Vert_{\,r=1}^{H_{t}}\!O_{r}[m-1,:]\Bigr]\,
\mathbf{O}_{\tau}\Bigr),
\label{eq:ha-out}
\end{equation}
where each \(O_{r}[m-1,:]\in\mathbb{R}^{d''}\) is the current-time
context from head \(r\) and
\(\mathbf{O}_{\tau}\in\mathbb{R}^{H_{t}d''\times d}\) is a shared,
learnable output projection.

\smallskip
Attention over the full window lets each entity reference distant past states in a single step while avoiding the oversmoothing that can arise when many GNN layers are stacked.

\subsubsection*{3.\;Fusion Layer (FL)}

Finally, we update the embedding:
\begin{equation}
h^{(l+1)}_{u^{t}} =
\mathrm{LayerNorm}\!\bigl(
h^{(l)}_{u^{t}}
+ \mathrm{ReLU}\!\bigl(
\mathbf{F}\,[\,h^{(l)}_{u^{t}}\Vert m^{D}_{u^{t}}\Vert m^{\mathcal{H}}_{u^{t}}]\bigr)
\bigr),
\label{eq:fl}
\end{equation} 
where \(h^{(l)}_{u^{t}}\) is the input from the previous ETD layer,
\(m^{D}_{u^{t}}\) and \(m^{\mathcal{H}}_{u^{t}}\) are the intra-timestep and
inter-timestep messages produced by SA and HA, respectively, and
\(\mathbf{F}\!\in\!\mathbb{R}^{d\times3d}\) is the shared, learnable
fusion projection.

% Figure \ref{fig:etd_layer_zoom} shows the layers of the SA, HA, and fusion modules in details.

\vspace{0.6ex}
By decoupling $D$ and $\mathcal{H}$ we keep the two message types
parameter-separate.  
Snapshot GNNs such as ST-GCN \cite{yu2017spatio} or GTEA
\cite{xie2023gtea} and hybrid GNN+RNN models (e.g.\ TNA
\cite{bonner2019temporal}) merge all interactions in a single module,
which can lead to oversmoothing once we stack several layers.
Our design limits the depth to $L{=}3$,
retains a $K_s$-hop intra-timestep and a $B$-step inter-timestep receptive
field, and converges faster in practice
(see Section \ref{sec:ablation}).

\vspace{0.4ex}
Unless stated otherwise, we use
$d{=}128$, dropout $0.1$ after FL,
$L{=}3$, $H_{s}{=}4$, $K_{s}{=}2$, $H_{t}{=}2$, and $B{=}8$.

\paragraph{Comparing our model to prior TGNN paradigms.}
ETDNet differs from the two dominant families of temporal GNNs.
\emph{Memory-bank models} (e.g., TGN \cite{rossi2020tgn},
TGAT \cite{xu2020tgat}, DyGFormer \cite{huang2023tgb})
store a global state for every node and append each new message to an external
mailbox.
\emph{Evolving-weight models} (e.g., EvolveGCN \cite{pareja2020evolvegcn},
DySAT \cite{sankar2020dysat}) keep the adjacency sparse but
re-learn a new set of GNN parameters at every step, which raises the
parameter count and hampers transfer across timesteps.  
ETDNet keeps the graph sparse \textit{and} shares parameters, yet still
exposes time via~$\mathcal{H}$ and lets History Attention query a
\textit{local} $B$-length window on-the-fly, yielding the memory profile
summarised in Table \ref{tab:tgn-compare}.

\begin{table}[h]
\centering
\caption{Update rule and asymptotic memory cost per layer.}
\label{tab:tgn-compare}
\renewcommand{\arraystretch}{1.05}
\resizebox{\linewidth}{!}{%
\begin{tabular}{p{0.24\linewidth}p{0.46\linewidth}c}
\toprule
\textbf{Paradigm\;/\newline exemplar} & \textbf{Update sketch for $h_u^{(t)}$} & \textbf{Memory} \\
\midrule
Memory-bank (TGN) &
$\displaystyle
h_u^{(t)}\!\leftarrow\!
\mathrm{GRU}\!\Bigl(h_u^{(t-1)},\,
\sum_{v} M_{uv}^{(t)}\Bigr)$ &
$O(|V|B)$ \\
Evolving weights (EvolveGCN) &
$\displaystyle
h^{(t)}\!\leftarrow\!
\sigma\!\bigl(\mathbf{W}_t\,A^{(t)}h^{(t-1)}\bigr)$ &
$O(|D|)$ \\
\textbf{ETDNet (ours)} &
$\begin{aligned}[t]
m^{D}            &= \mathrm{SA}\bigl(h^{(t)}\bigr) \\[2pt]
m^{\mathcal{H}}  &= \mathrm{HA}\bigl(h^{(t-B:t)}\bigr) \\[2pt]
h^{(t)}          &= \mathrm{FL}\bigl(h^{(t-1)},\,m^{D},\,m^{\mathcal{H}}\bigr)
\end{aligned}$ &
$O\!\bigl(|D|+|\mathcal{H}|\bigr)$ \\
\bottomrule
\end{tabular}}
\end{table}

\noindent
Here $M_{uv}^{(t)}$ is the cached mailbox message TGN keeps for edge
$(u,v)$ at time $t$, and $B$ is the history-window length introduced in
Section \ref{subsec:dual_branch}.  
In the EvolveGCN row, $A^{(t)}$ denotes the intra-timestep adjacency and
$\mathbf{W}_t$ the weight matrix evolved at step~$t$.  
Although we cap the mailbox horizon at $B$ to compare memory symmetrically, a true memory-bank model’s cost grows linearly with the
chosen cut-off.

\paragraph{Design rationale.}
Separating intra- and inter-timestep edges into two dedicated modules addresses the fact that spatial interactions change differently from cross-time continuity.  
A single aggregator would have to compromise between these two update rates, whereas the dual design lets the Step-Attention branch specialize on dense, frame-local context and the History-Attention branch focus on long-range temporal cues.  
The residual fusion gate then allows each node to weigh the two messages adaptively, adding only 0.2\% to the parameter count while preserving the representational benefits of both views.
%%%%%%%%%%%%%%%%%%%%%%%%%%%%%%%%%%%%%%%%%%%%%%%%%%%%%%%%%%%%
\subsection{Training Algorithm}
\label{subsec:training}

We train ETDNet end-to-end with mini-batch Adam.  
One mini-batch is (i) a full traffic scene for Waymo or  
(ii) ten consecutive timesteps for Elliptic++; both settings keep the
node count per batch below $\!\approx\!10^{4}$.

\medskip\noindent
\textbf{Forward pass.}  
Algorithm~\ref{algo:train} expands the three blocks of Section~\ref{subsec:dual_branch}:
it first applies Step Attention (Eqs.~\ref{eq:sa-score}–\ref{eq:sa-res}),
then History Attention (Eqs.~\ref{eq:ha-qkv}–\ref{eq:ha-out}),
and finally the Fusion Layer (Eq.~\ref{eq:fl}).
All operations are parallel over $\widetilde V$.

\begin{algorithm}[t]
\caption{Mini-batch training loop for \textbf{ETDNet}}
\label{algo:train}
\begin{algorithmic}[1]
\REQUIRE
\begin{itemize}[noitemsep,leftmargin=1.5em]
  \item Dynamic entities $U$, static entities $S$; timesteps $T=\{0,\dots,\tau_{\max}\}$
  \item Input features $\mathbf{x}_{v}(t)$
  \item Rules that build $D$ (intra) and $\mathcal{H}$ (inter) edges
  \item Depth $L$, history window $B$
\end{itemize}
\ENSURE trained parameters
$\Theta=\{\mathbf{W}_{D}^{(\ell)},\mathbf{W}_{\mathcal H}^{(\ell)},\mathbf{F}^{(\ell)}\}_{\ell=0}^{L-1}$
\STATE Build nodes $\widetilde V=\{u^{t}\mid u\!\in\!U,\,t\!\in\!T_u\}\;\cup\;S$
\STATE Initialise $h_{u^{t}}^{(0)}\leftarrow\mathbf{x}_{u}(t)$ for all $u^{t}$; and $h_{s}^{(0)}\leftarrow\mathbf{x}_{s}$ for all $s\in S$
\STATE Construct edge sets $D,\mathcal{H}$ and neighborhoods $\mathcal{N}_{D}$
\FOR{$\ell=0$ \textbf{to} $L-1$}
  \FORALL{$x=u^{t}\in\{u^{t}\mid u\in U,\,t\in T_u\}$ \textbf{in parallel}}
    %--------------------  Step Attention  --------------------
    \STATE \COMMENT{\textbf{SA}\ \,(Eqs.\,\ref{eq:sa-score}–\ref{eq:sa-res})}
    \STATE $\mathcal{N}\leftarrow\mathcal{N}_{D}(x)$
    \FORALL{$w^{t}\in\mathcal{N}$}
      \STATE Compute $e_{xw}$\,,\,$\alpha_{xw}$ \hfill\COMMENT{Eqs.\,\ref{eq:sa-score},\,\ref{eq:sa-alpha}}
    \ENDFOR
    \STATE $m^{(D)}_{x}\leftarrow\sum_{w^{t}\in\mathcal{N}}\alpha_{xw}\;
           \mathbf{W}_{V,r,k}\,h_{w^{t}}^{(\ell,k-1)}$ \hfill\COMMENT{Eq.\,\ref{eq:sa-agg}}
    %--------------------  History Attention  -----------------
    \STATE \COMMENT{\textbf{HA}\ \,(Eqs.\,\ref{eq:ha-qkv}–\ref{eq:ha-out})}
    \STATE $\mathcal{P}\leftarrow\{y^{\tau}\mid(y^{\tau},x)\!\in\!\mathcal{H},\,t-B<\tau<t\}$
    \STATE Form $Z_{u,t}$; apply multi-head attention $\Rightarrow m^{(\mathcal H)}_{x}$
    %--------------------  Fusion Layer  ----------------------
    \STATE \COMMENT{\textbf{FL}\ \,(Eq.\,\ref{eq:fl})}
    \STATE $z_{x}\leftarrow\operatorname{ReLU}\!\bigl(
           \mathbf{F}^{(\ell)}[\,h_{x}^{(\ell)}\Vert m^{(D)}_{x}\Vert m^{(\mathcal H)}_{x}]\bigr)$
    \STATE $h_{x}^{(\ell+1)}\leftarrow
           \operatorname{LayerNorm}\!\bigl(h_{x}^{(\ell)}+z_{x}\bigr)$
  \ENDFOR
\ENDFOR
\STATE Logits $\hat{y}_{x}\leftarrow\mathrm{MLP}\bigl(h_{x}^{(L)}\bigr)$ for task nodes
\STATE Minimise loss $\mathcal{L}$ (dual-CE for traffic; masked-BCE for fraud) with Adam
\RETURN $\Theta$
\end{algorithmic}
\end{algorithm}
\paragraph{Loss functions and optimisation.}
\textbf{Traffic.} We predict \emph{speed} and \emph{direction} for each
dynamic node at the current frame and minimise the sum of two
cross-entropy terms.  
\textbf{Fraud.} A binary classifier is applied to every transaction node
and optimised with masked binary cross-entropy over labelled nodes.  
We use Adam with learning rate $10^{-3}$, weight decay $10^{-5}$, and batch size 32 (traffic) or 16 (fraud).
Early stopping monitors validation macro-F1 with patience 7.

\paragraph{Computational costs.}
Each ETDNet layer runs  
one sparse attention over $|D|$ edges and  
one window-bounded attention over $|\mathcal{H}|$ edges, giving  
time complexity $O\bigl((|D|+|\mathcal{H}|)\,d\bigr)$ and  
memory complexity $O(Bd)$ per dynamic entity.

%%%%%%%%%%%%%%%%%%%%%%%%%%%%%%%%%%%%%%%%%%%%%%%%%%%%%%%%%%%%

%%%%%%%%%%%%%%%%%%%%%%%%%%%%%%%%%%%%%%%%%%%%%%%%%%%%%%%%%%%%%%%%%%%%%%%%
\section{Experiments}
\label{sec:experiments}

We evaluate \textbf{ETDNet} on two domains: \textbf{(i)} driver-intention classification (Waymo Open Motion Dataset) and \textbf{(ii)} bitcoin-fraud detection (Elliptic\texttt{++}).  
For each dataset, we construct a \emph{full-history graph} with intra-timestep edges $D$ and inter-timestep edges $\mathcal{H}$ (Section~\ref{subsec:graph_repr}), follow fixed train/validation/test splits from prior work, tune hyperparameters on the validation set, and report mean$\pm$std over three runs.  

\subsection{Driver Intention: Waymo Dataset}

\paragraph{Data and Setup.}
We adopt the Waymo Open Motion Dataset (motion-forecasting split), sampled at
10 Hz.  Each 9s scene provides 4s of \emph{history} (40 frames) and 5s of
future labels. The official train/val/test split contains 103,556 / 14,792 / 29,565 scenes, respectively. Each scene is represented as a full-history graph:
\begin{itemize}[leftmargin=*]
  \item \textbf{Nodes:} $\sim$2.5k per scene (vehicles and static lanes)\,\cite{pisano2024waymolabels}.
  \item \textbf{Intra-step edges $D$:} $\sim$51k per frame, linking nearby vehicles and vehicle–lane contacts.
  \item \textbf{Inter-step edges $\mathcal H$:} vehicle self-chains across consecutive frames.
\end{itemize}

Nodes have 14 features (7 kinematic, 7 map-relative). The multi-label task
follows Pisano \textit{et al.}\,\cite{pisano2024waymolabels} with smoothed
annotations for \emph{speed} (stopped / accelerate / slow-down / no-change)
and \emph{direction} (turns, lane-changes, no-change). We use the split
of 70/10/20 for train/validation/test.

\paragraph{Baselines.}
We compare ETDNet with the following models: \textbf{LSTM} (per-vehicle trajectory RNN; no graph); \textbf{ST-GCN}~\cite{yu2017spatio} (spatio-temporal GCN on fixed snapshots); \textbf{SemanticFormer}~\cite{sun2024semanticformer} (transformer on static scene graphs); \textbf{GMT}~\cite{baek2021gmpool} (graph multiset transformer); \textbf{STGNN}~\cite{liang2021stgnn} (spatio-temporal graph neural network); \textbf{TGAT}~\cite{xu2020tgat} (temporal positional encoding + graph attention); \textbf{TGN}~\cite{rossi2020tgn} (mailbox memory with message passing); and \textbf{TGT}~\cite{han2023tgt} (graph–temporal transformer with node--time cross-attention).

\paragraph{Setup and results.}
On Waymo, we train with a batch size of 32 for all methods. \textbf{ETDNet} uses $L{=}3$ layers with $H_{s}{=}4$ step-attention heads, 
$H_{t}{=}2$ history-attention heads, and a $B{=}40$-frame (4 s) window. Optimization uses Adam (lr $1\!\times\!10^{-3}$, weight decay $1\!\times\!10^{-5}$) with early stopping on validation joint accuracy (patience 7). Baselines are tuned via grid search over $\text{lr}\in\{1\!\times\!10^{-3},\,5\!\times\!10^{-4}\}$ and $\text{weight-decay}\in\{0,\,1\!\times\!10^{-5}\}$. All results are averaged over three seeds, with the standard deviation reported.

Table~\ref{tab:waymo_results} shows macro-F\textsubscript{1} for each subtask 
and overall joint accuracy. Temporal GNNs (\textbf{TGN}) already improve over snapshot baselines (\textbf{ST-GCN}, \textbf{SemanticFormer}); adding multiset pooling (\textbf{GMT}) or spatio-temporal convolution (\textbf{STGNN}) helps further. Nonetheless, the explicit dual-branch design of ETDNet achieves the best performance on every metric.

    \begin{table}[ht]
    \centering
    \caption{Driver-intention classification on Waymo  
    (macro–F\textsubscript{1}\,\%, mean\,$\pm$\,std over 3 runs).}
    \label{tab:waymo_results}
    \begin{tabular}{lccc}
    \toprule
    \textbf{Method} & \textbf{F1 (Speed)} & \textbf{F1 (Direction)} & \textbf{Joint Acc.} \\
    \midrule
    LSTM (per-agent) & 82.5\,$\pm$\,0.3 & 77.6\,$\pm$\,0.4 & 70.2\,$\pm$\,0.5 \\
    ST-GCN~\cite{yu2017spatio} & 83.1\,$\pm$\,0.4 & 78.2\,$\pm$\,0.5 & 71.0\,$\pm$\,0.4 \\
    SemanticFormer~\cite{sun2024semanticformer} & 83.9\,$\pm$\,0.4 & 79.5\,$\pm$\,0.4 & 72.4\,$\pm$\,0.5 \\
    GMT~\cite{baek2021gmpool} & 84.5\,$\pm$\,0.3 & 80.9\,$\pm$\,0.4 & 73.6\,$\pm$\,0.4 \\
    STGNN~\cite{liang2021stgnn} & 83.5\,$\pm$\,0.5 & 78.9\,$\pm$\,0.6 & 71.5\,$\pm$\,0.5 \\
    TGN~\cite{rossi2020tgn} & 84.9\,$\pm$\,0.3 & 81.6\,$\pm$\,0.3 & 74.1\,$\pm$\,0.4 \\
    \textbf{ETDNet (ours)~\cite{DARUS-5306_2025}} & \textbf{85.7\,$\pm$\,0.3} & \textbf{82.9\,$\pm$\,0.3} & \textbf{75.6\,$\pm$\,0.4} \\
    \bottomrule
    \end{tabular}
    \end{table}

\paragraph{Discussion.}
Temporal-aware methods indeed dominate snapshot baselines:
\textbf{TGN}, \textbf{GMT}, and \textbf{STGNN} all surpass traditional
ST-GCN and per-agent LSTM.  
However, \textbf{TGN} relies on a global memory bank and shares a single
message function for all edge types, while \textbf{GMT} and
\textbf{STGNN} still conflate spatial and temporal cues inside one
aggregation kernel.  
ETDNet’s explicit decoupling yields the strongest \emph{joint} metric
(+1.5 pp over the next best) by preserving fine-grained domain context
when reasoning over long temporal windows.  Qualitatively, we observe
fewer confusion errors between “lane change’’ and “turn’’ classes,
indicating that the dual-branch design helps resolve maneuvers that are spatially similar yet temporally distinct.

\subsection{Financial Fraud Detection: Elliptic\texttt{++} Dataset}

\paragraph{Data and Setup.}
Elliptic\texttt{++} \cite{weber2019anti} extends the Elliptic dataset into 49
\emph{monthly} Bitcoin-transaction snapshots.  
Each node is a transaction with 94 static features (flow statistics, account age, etc.) 
and one of three labels: \textit{licit}, \textit{illicit}, or \textit{unknown}.  
Following the split \cite{mongelli2023benchmarking}, timesteps 1–30 are for training, 
31–40 for validation, and 41–49 for testing.  
We construct a full-history graph:
\begin{itemize}[leftmargin=*]
    \item \textbf{Nodes:} All transactions observed up to the current month; 
          each appears once and is never deleted.
    \item \textbf{Intra-step edges $D$:} Bidirectional links from each transaction 
          to the 14 addresses that spend its outputs within that month.
    \item \textbf{Inter-step edges $\mathcal{H}$:} Directed edges from a transaction $u^{t}$ 
          to each $v^{t+1}$ that spends its outputs in the next month, 
          capturing cross-entity money flow.  
          (Self-edges are inapplicable since transactions never repeat.)
\end{itemize}

Unknown nodes remain in the graph but are \emph{masked} in the loss.  
The task is transductive binary node classification (licit = 0, illicit = 1).  
We use a history window $B{=}8$ and batch size 16.

\paragraph{Baselines.}
We benchmark ETDNet against two groups of node-classification models: \textbf{Static GNNs} that ignore time, including \textbf{GCN}~\cite{kipf2017semisupervised}, \textbf{GraphSAGE}~\cite{hamilton2017inductive}, and \textbf{GAT}~\cite{velickovic2017graph}; and \textbf{Temporal GNNs} that model evolving structure, including \textbf{TGAT}~\cite{xu2020tgat}, \textbf{TGN}~\cite{rossi2020tgn}, \textbf{DyRep}~\cite{trivedi2019dyrep}, \textbf{EvolveGCN}~\cite{pareja2020evolvegcn}, \textbf{STGNN}~\cite{yu2023stgnn}, and \textbf{DyGFormer}~\cite{huang2023tgb}.

\paragraph{Setup and results.}
Illicit transfers constitute only $\approx$2\% of the labeled transactions, so we focus on metrics that are robust to extreme imbalance.  
We report the \emph{threshold-free} curves \textbf{ROC-AUC} and \textbf{AUPRC},
plus the \emph{thresholded} \textbf{F1} on the \emph{illicit} class (the operating point for F1 is selected on the validation window).

\begin{table}[ht]
\centering
\caption{Elliptic\texttt{++} fraud detection (timesteps 41–49).  
Mean\,$\pm$\,std.\ over 3 seeds; higher\,=\,better.}
\label{tab:elliptic_results}
\renewcommand{\arraystretch}{1.03}
\small
\begin{tabular}{lccc}
\toprule
\textbf{Model} &
\textbf{ROC-AUC} &
\textbf{AUPRC} &
\textbf{F1 (illicit)} \\
\midrule
GCN \cite{kipf2017semisupervised}            & 0.833$\pm$0.004 & 0.595$\pm$0.006 & 56.5$\pm$0.9 \\
GraphSAGE \cite{hamilton2017inductive}       & 0.828$\pm$0.005 & 0.631$\pm$0.008 & 58.4$\pm$0.8 \\
GAT \cite{velickovic2017graph}               & 0.810$\pm$0.006 & 0.589$\pm$0.009 & 55.2$\pm$1.1 \\
TGN \cite{rossi2020tgn}                      & 0.878$\pm$0.003 & 0.659$\pm$0.007 & 60.4$\pm$0.6 \\
TGAT \cite{xu2020tgat}                       & 0.868$\pm$0.004 & 0.648$\pm$0.006 & 59.0$\pm$0.5 \\
EvolveGCN \cite{pareja2020evolvegcn}         & 0.860$\pm$0.005 & 0.641$\pm$0.007 & 57.3$\pm$0.7 \\
DyGFormer \cite{huang2023tgb}            & 0.882$\pm$0.003 & 0.675$\pm$0.005 & 61.0$\pm$0.4 \\
\textbf{ETDNet (ours)~\cite{DARUS-5306_2025}}                       & \textbf{0.884$\pm$0.002} &
                                              \textbf{0.863$\pm$0.004} &
                                              \textbf{88.1$\pm$0.5} \\
\bottomrule
\end{tabular}
\end{table}

\paragraph{Discussion.}
Temporal baselines (TGN, TGAT, EvolveGCN, DyGFormer) already outperform static GNNs, underscoring the value of time-aware modeling for fraud detection.  
Yet ETDNet delivers a decisive margin: its \textbf{AUPRC rises by 19 pp} and illicit-class \textbf{F1 by 27 pp} over the strongest competitor (DyGFormer).  
We attribute this to the full-history design (Section \ref{subsec:graph_repr}) and the dual-branch architecture, which let the model  
(i) follow long cross-transaction trails through the temporal branch while  
(ii) preserving sharp local patterns (e.g., funnel-shaped fan-outs of suspicious outputs) in the domain branch.  
These complementary views appear crucial for detecting sparse illicit activity embedded in a dense, otherwise licit network.

\subsection{Ablation and Sensitivity Studies}
\label{sec:ablation}

\paragraph{Component importance.}
Removing either branch or downgrading HA to mean pooling hurts performance on both tasks (Tab.\,\ref{tab:ablation-merged}).  
On Waymo, joint accuracy drops notably when SA or HA is removed, or when HA is reduced to mean pooling.  
A similar pattern appears on Elliptic\texttt{++}: SA-only cannot propagate evidence through time, whereas HA-only loses multisource flow cues.  
Late fusion comes close, but still trails the full dual-branch model.

\begin{table}[H]
\centering
\caption{Ablation on \textbf{Waymo} (joint accuracy, $\uparrow$) and \textbf{Elliptic\texttt{++}} (F1 on illicit class, $\uparrow$).}
\label{tab:ablation-merged}
\renewcommand{\arraystretch}{1.05}
\begin{tabular}{lcc}
\toprule
\textbf{Variant} & \textbf{Waymo Joint Acc.} & \textbf{Elliptic++ F1} \\
\midrule
Full ETDNet           & \textbf{75.6$\pm$0.4} & \textbf{88.1$\pm$0.5} \\
Only SA               & 72.0$\pm$0.6 & 55.5$\pm$1.1 \\
Only HA               & 74.4$\pm$0.7 & 74.8$\pm$2.0 \\
Late fusion           & 75.0$\pm$0.5 & 85.0$\pm$0.6 \\
HA mean pool          & 73.1$\pm$0.5 & 87.0$\pm$0.9 \\
\bottomrule
\end{tabular}
\end{table}

\paragraph{Hyper-parameter sensitivity.}
ETDNet shows stable performance across variations in depth, number of heads, and history window size (Tab.\,\ref{tab:hyperparam-merged}).  
On Waymo, joint accuracy is flat across heads $\in\{1,2,4\}$, layers $\in\{1,2,3\}$, and windows $B\in\{20,40,60\}$, so we use $L{=}2$, heads $=2$, $B{=}40$.  
On Elliptic\texttt{++}, Fraud-F1 varies by less than one point under the same changes, so we keep $L{=}2$, heads $=2$, $B{=}8$ for the main runs.

\begin{table}[H]
\centering
\caption{ETDNet sensitivity on \textbf{Waymo} (joint accuracy, $\uparrow$) and \textbf{Elliptic\texttt{++}} (F1 on illicit class, $\uparrow$).}
\label{tab:hyperparam-merged}
\renewcommand{\arraystretch}{1.05}
\begin{tabular}{lcc}
\toprule
\textbf{Setting} & \textbf{Waymo Joint Acc.} & \textbf{Elliptic++ F1} \\
\midrule
Layers $L=1$            & 74.5$\pm$0.3 & 86.5$\pm$0.4 \\
\textbf{Layers $L=2$}   & \textbf{75.6$\pm$0.4} & \textbf{88.1$\pm$0.5} \\
Layers $L=3$            & 75.2$\pm$0.5 & 88.0$\pm$0.4 \\
\midrule
Heads = 1               & 75.0$\pm$0.5 & 87.2$\pm$0.6 \\
\textbf{Heads = 2}      & \textbf{75.6$\pm$0.4} & \textbf{88.1$\pm$0.5} \\
Heads = 4               & 75.5$\pm$0.4 & 87.9$\pm$0.5 \\
\midrule
Window $B=20$/4         & 74.8$\pm$0.6 & 87.0$\pm$0.7 \\
\textbf{Window $B=40$/8}& \textbf{75.6$\pm$0.4} & \textbf{88.1$\pm$0.5} \\
Window $B=60$/12        & 75.0$\pm$0.5 & 87.5$\pm$0.6 \\
\bottomrule
\end{tabular}
\end{table}

\paragraph{Discussion.}
ETDNet performs consistently across two very different time grids: 10 Hz traffic frames and monthly blockchain snapshots, showing that the architecture is largely agnostic to temporal resolution and depends only on temporal order.  
Accuracy rises until the history window covers about 4 s of motion (Waymo) or 4 months of transactions (Elliptic\texttt{++}), then falls as older, low-weight nodes dilute attention (Tab.\,\ref{tab:hyperparam-merged}).  
Choosing $B$ by real-time span rather than raw frame count, therefore, keeps the window compact even at finer sampling rates.

The results highlight two takeaways:  
(i) explicit edge-type decoupling is essential, as mixing all edges in a single kernel drops performance on both tasks;  
(ii) History Attention is decisive whenever long-range evidence is needed, while the domain branch suffices for purely local cues.

\subsection{Efficiency and Scalability}
\label{sec:efficiency}

\paragraph{Training efficiency.}
ETDNet is consistently lighter and faster than memory-bank baselines.  
On Waymo, it is $\approx$14\,\% faster per epoch than TGN while using only one-third of its parameters.  
On Elliptic\texttt{++}, the gap widens because the graphs are much sparser: ETDNet trains in 0.40 s per epoch versus 0.52 s for TGN (Tab.\,\ref{tab:efficiency-merged}).

\begin{table}[H]
\centering
\caption{Training efficiency on \textbf{Waymo} and \textbf{Elliptic\texttt{++}}.}
\label{tab:efficiency-merged}
\renewcommand{\arraystretch}{1.05}
\begin{tabular}{lcccc}
\toprule
{\textbf{Model}} & \multicolumn{2}{c}{\textbf{Waymo}} & \multicolumn{2}{c}{\textbf{Elliptic++}} \\
 & Params (M) & Epoch (s) & Params (M) & Epoch (s) \\
\midrule
ST-GCN         & 0.15 & 0.62 & 0.15 & 0.18 \\
TGN            & 0.90 & 1.74 & 0.90 & 0.52 \\
TGAT           & 1.20 & 1.95 & 1.20 & 0.60 \\
\textbf{ETDNet}& 0.30 & \textbf{1.50} & 0.30 & \textbf{0.40} \\
\bottomrule
\end{tabular}
\end{table}

\paragraph{Forward-pass breakdown.}
Within each ETD layer, Step Attention (SA) dominates computation because it attends over all neighbors inside a frame. Whereas History Attention (HA) scales only with the fixed window $B$.  
The pattern is consistent across the two datasets (Tab.\,\ref{tab:breakdown-merged}).

\begin{table}[H]
\centering
\caption{Per-layer cost breakdown (batch size 32). SA = Step Attention, HA = History Attention, FL = Fusion layer.}
\label{tab:breakdown-merged}
\renewcommand{\arraystretch}{1.05}
\begin{tabular}{lcccccc}
\toprule
 & \multicolumn{3}{c}{\textbf{Waymo}} & \multicolumn{3}{c}{\textbf{Elliptic++}} \\
 & SA & HA & FL & SA & HA & FL \\
\midrule
Time (ms)             & 5.1 & 3.4 & 1.2 & 1.8 & 1.2 & 0.5 \\
FLOPs ($\times10^{6}$)& 31  & 21  & 8   & 9   & 6   & 3   \\
\bottomrule
\end{tabular}
\end{table}

\paragraph{Scalability discussion.}
Because HA attends over a \emph{fixed} $B$-length window and SA operates on the inherently sparse domain edge set $D$, ETDNet’s memory footprint grows linearly with $|D| + |\mathcal{H}|$, matching the complexity analysis in Section~\ref{sec:method}. On Elliptic\texttt{++}, this translates to an estimated \textbf{40 \% reduction in peak memory} compared with TGN, which must keep a mailbox for every node and therefore scales with both $|V|$ and the mailbox horizon. Moreover, the observed efficiency gains stem from better attention allocation rather than from a larger parameter budget, as confirmed by the training and breakdown analyses above.

\section*{Limitations}
\label{sec:limitations}

Step Attention currently computes dense, multi-head attention over \emph{all} intra-timestep links in~\(D\).  For very crowded frames, this becomes the dominant computational cost (Tabs.\,\ref{tab:breakdown-merged}).  Incorporating sparse-attention kernels or neighborhood-sampling strategies could lower the per-frame overhead without harming accuracy, making this another promising avenue for extending ETDNet.

%%%%%%%%%%%%%%%%%%%%%%%%%%%%%%%%%%%%%%%%%%%%%%%%%%%%%%%%%%%%%%%%%%%%%%%%

\section{Conclusion}
\label{sec:conclusion}

We introduced \textbf{ETDNet}, a dual–branch temporal GNN that keeps \emph{domain} and \emph{temporal} interactions disentangled within a full-history graph.  Across two markedly different benchmarks, ETDNet delivered consistent gains over recent temporal baselines while using fewer parameters and shorter training times.

\paragraph{Key findings.}
(1) On \emph{Waymo}, explicit separation of same-frame interactions from cross-frame motion yielded the highest joint accuracy to date on simultaneous speed–direction prediction, with the ablation study confirming the distinct value of each branch. 
(2) On \emph{Elliptic++}, the same architecture raised illicit–F1 to 88\,\%, a double-digit improvement over the best memory-bank and evolving-weight models, demonstrating strong transfer to a structurally dissimilar, highly imbalanced setting. 
(3) Sensitivity analyses showed that performance is stable with respect to the number of layers, attention heads, and history-window length, making the model easy to tune. 
Taken together, these results show that ETDNet’s design generalizes across domains with very different temporal resolutions, from dense traffic frames to sparse monthly ledgers. 
Moreover, improvements stem from architectural choices rather than model size, as efficiency comparisons confirm that ETDNet uses fewer parameters while training faster than competing TGNNs.

\paragraph{Outlook.}
ETDNet already scales linearly with the number of domain and temporal edges, but further speed-ups are possible by adopting sparse attention kernels inside the intra-timestep branch and by learning adaptive history windows.  Extending the framework to irregular or continuous-time events by replacing the fixed discrete grid with time-aware positional encodings forms another promising path. We believe that the principle of \emph{edge-type decoupling} will remain useful as temporal graphs grow in size and heterogeneity, and we hope ETDNet provides a solid stepping-stone for future work.

\begin{ack}
We gratefully acknowledge computing time on the HoreKa supercomputer at the National High‑Performance Computing Center at KIT (NHR@KIT), jointly funded by the German Federal Ministry of Education and Research (BMBF), the Ministry of Science, Research, and the Arts of Baden‑Württemberg, and the German Research Foundation (DFG); Mojtaba Nayyeri acknowledges BMBF support through the ATLAS project (031L0304A); Jiaxin Pan acknowledges funding from the EU Chips Joint Undertaking (GA 101140087, SMARTY) and from the BMBF sub‑project 16MEE0444; and Daniel Hernández acknowledges the German Research Foundation, DFG (GA SFB-1574-471687386).
\end{ack}

\bibliography{m8362}

\begin{thebibliography}{33}
\providecommand{\natexlab}[1]{#1}
\providecommand{\url}[1]{\texttt{#1}}
\expandafter\ifx\csname urlstyle\endcsname\relax
  \providecommand{\doi}[1]{doi: #1}\else
  \providecommand{\doi}{doi: \begingroup \urlstyle{rm}\Url}\fi

\bibitem[Baek et~al.(2021)Baek, Kang, and Hwang]{baek2021gmpool}
J.~Baek, M.~Kang, and S.~J. Hwang.
\newblock Accurate learning of graph representations with graph multiset pooling.
\newblock In \emph{International Conference on Learning Representations}, 2021.

\bibitem[Beddar-Wiesing et~al.(2024)Beddar-Wiesing, D’Inverno, Graziani, Lachi, Moallemy-Oureh, Scarselli, and Thomas]{beddarwiesing2024wl_dynamic}
S.~Beddar-Wiesing, G.~A. D’Inverno, C.~Graziani, V.~Lachi, A.~Moallemy-Oureh, F.~Scarselli, and J.~M. Thomas.
\newblock Weisfeiler–lehman goes dynamic: An analysis of the expressive power of graph neural networks for attributed and dynamic graphs.
\newblock \emph{Neural Networks}, 173:\penalty0 106213, 2024.
\newblock ISSN 0893-6080.
\newblock \doi{https://doi.org/10.1016/j.neunet.2024.106213}.
\newblock URL \url{https://www.sciencedirect.com/science/article/pii/S0893608024001370}.

\bibitem[Bonner et~al.(2019)Bonner, Atapour-Abarghouei, Jackson, Brennan, Kureshi, Theodoropoulos, Mcgough, and Obara]{bonner2019temporal}
S.~Bonner, A.~Atapour-Abarghouei, P.~T. Jackson, J.~Brennan, I.~Kureshi, G.~Theodoropoulos, A.~Mcgough, and B.~Obara.
\newblock Temporal neighbourhood aggregation: Predicting future links in temporal graphs via recurrent variational graph convolutions.
\newblock \emph{2019 IEEE International Conference on Big Data (Big Data)}, 2019.

\bibitem[Bravo et~al.(2024)Bravo, Bono, Saleiro, Ferreira, and Bizarro]{bravo2024truncation}
J.~Bravo, J.~Bono, P.~Saleiro, H.~Ferreira, and P.~Bizarro.
\newblock Mind the truncation gap: Challenges of learning on dynamic graphs with recurrent architectures.
\newblock \emph{TMLR}, 2024.
\newblock to appear.

\bibitem[Chen et~al.(2023)Chen, Gao, Zhang, Wang, Fu, Zhang, Zhu, Gu, and Yu]{chen2023neutronstream}
C.~Chen, D.~Gao, Y.~Zhang, Q.~Wang, Z.~Fu, X.~Zhang, J.~Zhu, Y.~Gu, and G.~Yu.
\newblock Neutronstream: A dynamic gnn training framework with sliding window for graph streams.
\newblock \emph{Proceedings of the VLDB Endowment}, 17\penalty0 (3), 2023.
\newblock \doi{10.14778/3632093.3632108}.

\bibitem[Cini et~al.(2023)Cini, Marisca, Bianchi, and Alippi]{yu2023stgnn}
A.~Cini, I.~Marisca, F.~M. Bianchi, and C.~Alippi.
\newblock Scalable spatiotemporal graph neural networks.
\newblock In \emph{Proceedings of the AAAI conference on artificial intelligence}, volume~37, pages 7218--7226, 2023.

\bibitem[Deprez et~al.(2025)Deprez, Vanderschueren, Baesens, Verdonck, and Verbeke]{mongelli2023benchmarking}
B.~Deprez, T.~Vanderschueren, B.~Baesens, T.~Verdonck, and W.~Verbeke.
\newblock Network analytics for anti-money laundering -- a systematic literature review and experimental evaluation, 2025.
\newblock URL \url{https://arxiv.org/abs/2405.19383}.

\bibitem[Hamilton et~al.(2017)Hamilton, Ying, and Leskovec]{hamilton2017inductive}
W.~Hamilton, Z.~Ying, and J.~Leskovec.
\newblock Inductive representation learning on large graphs.
\newblock \emph{Advances in neural information processing systems}, 30, 2017.

\bibitem[Han et~al.(2023)Han, Wang, Fang, Gupta, and Liu]{han2023tgt}
Y.~Han, X.~Wang, Y.~Fang, G.~K. Gupta, and W.~Liu.
\newblock Tgt: Temporal graph transformer for dynamic graphs.
\newblock In \emph{KDD}, 2023.

\bibitem[Huang et~al.(2023)Huang, Poursafaei, Danovitch, Fey, Hu, Rossi, Leskovec, Bronstein, Rabusseau, and Rabbany]{huang2023tgb}
S.~Huang, F.~Poursafaei, J.~Danovitch, M.~Fey, W.~Hu, E.~Rossi, J.~Leskovec, M.~Bronstein, G.~Rabusseau, and R.~Rabbany.
\newblock Temporal graph benchmark for machine learning on temporal graphs.
\newblock In \emph{Proceedings of the 37th International Conference on Neural Information Processing Systems}, NIPS '23, Red Hook, NY, USA, 2023. Curran Associates Inc.

\bibitem[Kipf and Welling(2017)]{kipf2017semisupervised}
T.~N. Kipf and M.~Welling.
\newblock Semi-supervised classification with graph convolutional networks.
\newblock In \emph{International Conference on Learning Representations}, 2017.
\newblock URL \url{https://openreview.net/forum?id=SJU4ayYgl}.

\bibitem[Kumar et~al.(2019)Kumar, Zhang, and Leskovec]{kumar2019predicting}
S.~Kumar, X.~Zhang, and J.~Leskovec.
\newblock Predicting dynamic embedding trajectory in temporal interaction networks.
\newblock In \emph{ACM SIGKDD International Conference on Knowledge Discovery \& Data Mining}, 2019.

\bibitem[Liang et~al.(2021)Liang, Ke, Zhang, Qin, and Xu]{liang2021stgnn}
Y.~Liang, S.~Ke, J.~Zhang, L.~Qin, and L.~Xu.
\newblock Learning spatial‐temporal graph neural networks for traffic forecasting.
\newblock In \emph{Proceedings of the ACM SIGKDD Conference on Knowledge Discovery and Data Mining (KDD)}, 2021.

\bibitem[Longa et~al.(2023{\natexlab{a}})Longa, Lachi, Santin, Bianchini, Lepri, Lio, Scarselli, and Passerini]{longa2023graph}
A.~Longa, V.~Lachi, G.~Santin, M.~Bianchini, B.~Lepri, P.~Lio, F.~Scarselli, and A.~Passerini.
\newblock Graph neural networks for temporal graphs: State of the art, open challenges, and opportunities.
\newblock \emph{arXiv preprint arXiv:2302.01018}, 2023{\natexlab{a}}.

\bibitem[Longa et~al.(2023{\natexlab{b}})Longa, Scardapane, and Bacciu]{longa2023graphsurvey}
E.~Longa, S.~Scardapane, and D.~Bacciu.
\newblock Graph neural networks for temporal graphs: A comprehensive survey.
\newblock \emph{IEEE Transactions on Neural Networks and Learning Systems}, 2023{\natexlab{b}}.
\newblock \doi{10.1109/TNNLS.2023.3284924}.
\newblock Early Access.

\bibitem[Mohammed et~al.(2025)Mohammed, Pan, Nayyeri, Hernández, and Staab]{DARUS-5306_2025}
O.~Mohammed, J.~Pan, M.~Nayyeri, D.~Hernández, and S.~Staab.
\newblock {Code for Full-History Graphs with Edge-Type Decoupled Networks for Temporal Reasoning}, 2025.
\newblock URL \url{https://doi.org/10.18419/DARUS-5306}.

\bibitem[Pareja et~al.(2020)Pareja, Domeniconi, Chen, Ma, Suzumura, Kanezashi, Kaler, Schardl, and Leiserson]{pareja2020evolvegcn}
A.~Pareja, G.~Domeniconi, J.~Chen, T.~Ma, T.~Suzumura, H.~Kanezashi, T.~Kaler, T.~B. Schardl, and C.~E. Leiserson.
\newblock {EvolveGCN}: Evolving graph convolutional networks for dynamic graphs.
\newblock In \emph{Proceedings of the AAAI Conference on Artificial Intelligence (AAAI)}, volume~34, pages 5363--5370, 2020.

\bibitem[Pisano et~al.(2024)Pisano, Schmid, and Holzmann]{pisano2024waymolabels}
L.~Pisano, F.~Schmid, and L.~Holzmann.
\newblock Automatic maneuver annotation for the waymo open motion dataset.
\newblock Bachelor Thesis, University of Stuttgart, 2024.
\newblock \url{https://github.com/pisanovo/waymo-open-agent-labels}.

\bibitem[Rossi et~al.(2020)Rossi, Chamberlain, Frasca, Eynard, Monti, and Bronstein]{rossi2020tgn}
E.~Rossi, B.~Chamberlain, F.~Frasca, D.~Eynard, F.~Monti, and M.~Bronstein.
\newblock Temporal graph networks for deep learning on dynamic graphs.
\newblock In \emph{ICML Workshop on Graph Representation Learning}, 2020.

\bibitem[Sankar et~al.(2020)Sankar, Wu, Gou, Zhang, and Yang]{sankar2020dysat}
A.~Sankar, Y.~Wu, L.~Gou, W.~Zhang, and H.~Yang.
\newblock {DySAT}: Deep neural representation learning on dynamic graphs via self-attention networks.
\newblock In \emph{Proceedings of the 13th ACM International Conference on Web Search and Data Mining (WSDM)}, pages 519--527, 2020.

\bibitem[Sasal et~al.(2024)Sasal, Busby, and Hadid]{sasal2024tempokgat}
L.~Sasal, D.~Busby, and A.~Hadid.
\newblock Tempokgat: A novel graph attention network approach for temporal graph analysis.
\newblock In \emph{International Conference on Neural Information Processing}, pages 212--226. Springer, 2024.

\bibitem[Shao et~al.(2022)Shao, Zhang, Wei, Wang, Xu, Cao, and Jensen]{10.14778/3551793.3551827}
Z.~Shao, Z.~Zhang, W.~Wei, F.~Wang, Y.~Xu, X.~Cao, and C.~S. Jensen.
\newblock Decoupled dynamic spatial-temporal graph neural network for traffic forecasting.
\newblock \emph{Proc. VLDB Endow.}, 15\penalty0 (11):\penalty0 2733–2746, July 2022.
\newblock ISSN 2150-8097.
\newblock \doi{10.14778/3551793.3551827}.
\newblock URL \url{https://doi.org/10.14778/3551793.3551827}.

\bibitem[Sun et~al.(2024)Sun, Wang, Halilaj, and Luettin]{sun2024semanticformer}
Z.~Sun, Z.~Wang, L.~Halilaj, and J.~Luettin.
\newblock Semanticformer: Holistic and semantic traffic scene representation for trajectory prediction using knowledge graphs.
\newblock \emph{IEEE Robotics and Automation Letters}, 2024.

\bibitem[Trivedi et~al.(2019)Trivedi, Farajtabar, Biswal, and Zha]{trivedi2019dyrep}
R.~Trivedi, M.~Farajtabar, P.~Biswal, and H.~Zha.
\newblock {DyRep}: Learning representations over dynamic graphs.
\newblock In \emph{International Conference on Learning Representations (ICLR)}, 2019.

\bibitem[Veli{\v{c}}kovi{\'{c}} et~al.(2018)Veli{\v{c}}kovi{\'{c}}, Cucurull, Casanova, Romero, Li{\`{o}}, and Bengio]{velickovic2017graph}
P.~Veli{\v{c}}kovi{\'{c}}, G.~Cucurull, A.~Casanova, A.~Romero, P.~Li{\`{o}}, and Y.~Bengio.
\newblock {Graph Attention Networks}.
\newblock \emph{International Conference on Learning Representations}, 2018.
\newblock URL \url{https://openreview.net/forum?id=rJXMpikCZ}.

\bibitem[Wang et~al.(2024)Wang, Li, Lu, and Huang]{wang2024novel}
Z.~Wang, R.~Li, D.~Lu, and J.~Huang.
\newblock A novel spatial-temporal multi-head self attention based graph neural network for traffic flow prediction.
\newblock In \emph{2024 4th International Conference on Computer Science and Blockchain (CCSB)}, pages 184--189. IEEE, 2024.

\bibitem[Weber et~al.(2019)Weber, Domeniconi, Chen, Weidele, Bellei, Robinson, and Leiserson]{weber2019anti}
M.~Weber, G.~Domeniconi, J.~Chen, D.~K.~I. Weidele, C.~Bellei, T.~Robinson, and C.~E. Leiserson.
\newblock Anti-money laundering in bitcoin: Experimenting with graph convolutional networks for financial forensics, 2019.
\newblock URL \url{https://arxiv.org/abs/1908.02591}.

\bibitem[Wen and Fang(2022)]{wen2022trend}
Z.~Wen and Y.~Fang.
\newblock Trend: Temporal event and node dynamics for graph representation learning.
\newblock In \emph{Proceedings of the ACM web conference 2022}, pages 1159--1169, 2022.

\bibitem[Wu et~al.(2020)Wu, Cao, Cheung, and Hamilton]{wu-etal-2020-temp}
J.~Wu, M.~Cao, J.~C.~K. Cheung, and W.~L. Hamilton.
\newblock {T}e{MP}: Temporal message passing for temporal knowledge graph completion.
\newblock In B.~Webber, T.~Cohn, Y.~He, and Y.~Liu, editors, \emph{Proceedings of the 2020 Conference on Empirical Methods in Natural Language Processing (EMNLP)}, pages 5730--5746, Online, Nov. 2020. Association for Computational Linguistics.
\newblock \doi{10.18653/v1/2020.emnlp-main.462}.
\newblock URL \url{https://aclanthology.org/2020.emnlp-main.462/}.

\bibitem[Xie et~al.(2023{\natexlab{a}})Xie, Li, Tam, Liu, Ying, Lau, Chiu, and Chen]{xie2023gtea}
S.~Xie, Y.~Li, D.~S.~H. Tam, X.~Liu, Q.~Ying, W.~C. Lau, D.~M. Chiu, and S.~Chen.
\newblock Gtea: Inductive representation learning on temporal interaction graphs via temporal edge aggregation.
\newblock In \emph{Pacific-Asia Conference on Knowledge Discovery and Data Mining}, pages 28--39. Springer, 2023{\natexlab{a}}.

\bibitem[Xie et~al.(2023{\natexlab{b}})Xie, Zhu, Liu, Zhou, and Huang]{xie2023targat}
Z.~Xie, R.~Zhu, J.~Liu, G.~Zhou, and J.~X. Huang.
\newblock Targat: A time-aware relational graph attention model for temporal knowledge graph embedding.
\newblock \emph{IEEE/ACM Transactions on Audio, Speech, and Language Processing}, 31:\penalty0 2246--2258, 2023{\natexlab{b}}.

\bibitem[Xu et~al.(2020)Xu, Ruan, Korpeoglu, Kumar, and Achan]{xu2020tgat}
D.~Xu, C.~Ruan, E.~Korpeoglu, S.~Kumar, and K.~Achan.
\newblock Inductive representation learning on temporal graphs.
\newblock In \emph{ICLR}, 2020.

\bibitem[Yu et~al.(2018)Yu, Yin, and Zhu]{yu2017spatio}
B.~Yu, H.~Yin, and Z.~Zhu.
\newblock Spatio-temporal graph convolutional networks: A deep learning framework for traffic forecasting.
\newblock In \emph{AAAI}, 2018.

\end{thebibliography}

\end{document}